\documentclass{article} 
\usepackage{iclr2026_conference,times}

\usepackage{amsmath,amsfonts,bm}

\def\eqref#1{equation~\ref{#1}}

\def\1{\bm{1}}

\DeclareMathAlphabet{\mathsfit}{\encodingdefault}{\sfdefault}{m}{sl}
\SetMathAlphabet{\mathsfit}{bold}{\encodingdefault}{\sfdefault}{bx}{n}

\usepackage{hyperref}
\usepackage{url}

\usepackage{subcaption}
\usepackage{amssymb}
\usepackage{accents} 
\usepackage{float}
\usepackage[utf8]{inputenc}
\usepackage{amsmath}
\usepackage{amsthm}
\usepackage{booktabs}
\usepackage[switch]{lineno}

\usepackage{soul}
\usepackage{url}
\usepackage{graphicx}
\usepackage{amsmath}
\usepackage{amsthm}
\usepackage{booktabs}
\usepackage{algorithm}
\usepackage{algorithmic}
\usepackage[switch]{lineno}

\title{Explaining How Quantization Disparately Skews a Model}

\author{Abhimanyu Bellam \& Jung-Eun Kim\thanks{Corresponding author.} \\
Department of Computer Science\\
North Carolina State University\\
Raleigh, NC, USA \\
}

\iclrfinalcopy 
\begin{document}

\maketitle

\begin{abstract}
    Post Training Quantization (PTQ) is widely adopted due to its high compression capacity and speed with minimal impact on accuracy. However, we observed that disparate impacts are exacerbated by quantization, especially for minority groups. Our analysis explains that in the course of quantization there is a chain of factors attributed to a disparate impact across groups during forward and backward passes. We explore how the changes in weights and activations induced by quantization cause cascaded impacts in the network, resulting in logits with lower variance, increased loss, and compromised group accuracies. We extend our study to verify the influence of these impacts on group gradient norms and eigenvalues of the Hessian matrix, providing insights into the state of the network from an optimization point of view. To mitigate these effects, we propose integrating mixed precision Quantization Aware Training (QAT) with dataset sampling methods and weighted loss functions, therefore providing fair deployment of quantized neural networks.
\end{abstract}



\section{Introduction}

With the onset of edge devices running deep neural networks for various tasks ranging across several domains, the demand for faster computation and model lightness has become more pronounced. To aid this, compression methods such as pruning \cite{han2015deep} and quantization \cite{hubara2016quantized} have taken the lead, producing little to no loss of accuracy with considerable memory and speed gains. Nevertheless, these methods do not account for the possible disparate impact they cause, and have been shown to have adverse effects on minority groups and exacerbate the shortcomings of their dense, counterpart model \cite{hooker2019compressed}. 


 In a streamline of model compression, \cite{tran2022pruning} recognized that magnitude pruning can exacerbate unfairness among classes. While pruning and quantization share a common objective of compressing a model, they are different in their approaches. Pruning involves removing weights or components that are insignificant according to a defined criterion. Whereas, quantization focuses on reducing the precision of the bits used to represent the weights and activations of the neural network. Notably, \cite{kuzmin2023pruning} demonstrated that quantization outperforms pruning-based strategies when similar model sizes and resource footprints are considered.  Furthermore, quantization is prominent for Large Language Models (LLMs) due to their large parameter sizes and requirement for reduced energy consumption \cite{kim2023memoryefficient,frantar2022gptq,pmlr-v202-dettmers23a}. 

\begin{figure*} [t]
    \centering
    \includegraphics[width=1.0\linewidth]{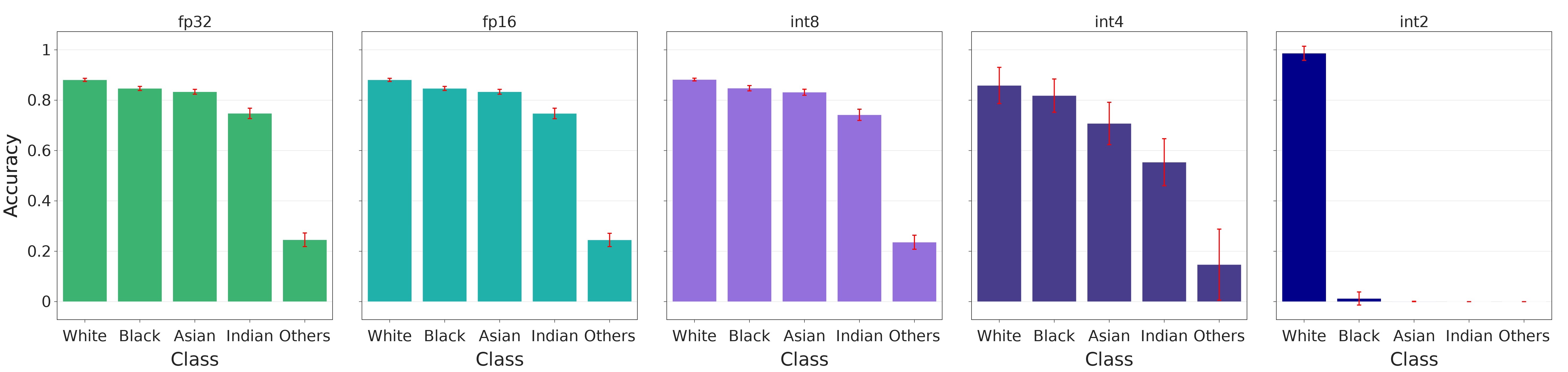}
    \caption{Accuracy for groups according to different quantization precisions on UTKFace dataset.}
    \label{fig:base_accuracies}
\end{figure*}

 We observed that quantization can exacerbate disparity of a model, especially for the minority group, as we show in Fig.~\ref{fig:base_accuracies}. The leftmost chart is pre-quantization. As the precision is reduced, the disparity is exacerbated further. When the model is quantized to $\mathtt{int2}$, the disparity is extreme. In this paper, we identify the factors that impact the disparity and optimization state in the forward and backward passes, respectively.

Post Training Quantization (PTQ) modifies the weights of the network while setting several weights to absolute zeros, thereby inducing sparsity, which together brings in disparate impacts of a model. Consequently, the logits suffer from a reduction in variance, similar to using high temperature scaling, while undergoing magnitude changes that lead to misclassifications. These factors finally alter the softmax probabilities and skew their distributions closer to the decision boundary towards low confidence regions, causing higher loss and group disparity. Additionally, PTQ shifts the model to a worse position in the optimization space, with larger gradient norms and eigenvalues of the Hessian matrix for minority classes, implying a potential for further optimization.


To combat these problems, we leverage the simplicity of dataset sampling methods to overcome the dataset imbalance, combine it with a weighted cross-entropy loss function to deal with example difficulty, and use mixed-precision Quantization Aware Training (QAT) to withstand the degradation of model performance due to low precision representations.




Our contributions are summarized as:
\begin{enumerate}
    \item We observed and showed that quantization can exacerbate disparity of a model, especially for minority groups.
    \item We identified the factors of PTQ that cause disparity: change in weights, increased sparsity, changes in logits, and reduced variance of softmax probabilities. These are cascaded factors in the forward pass. 
    \item We examine the degradation caused to the model's state in the optimization space, using gradient norms and eigenvalues of the Hessian matrix of a quantized model.
    \item We proposed a mitigating quantization approach that incorporates sampling methods and weighted loss functions for improved fairness.
\end{enumerate}

A preliminary version of this work is published in \cite{FITML2024Bellam}.




\section{Related Work}



\subsection{Quantization} 
This competent method contributes to a significant body of work in model compression with the introduction of compatible hardware, giving rise to diverse methodologies such as PTQ and QAT. PTQ is a domain where quantization is performed after the model is fully trained on a dataset without further retraining \cite{banner2019posttraining,zhao2019improving,choukroun2019lowbit,pmlr-v119-wang20c,li2021brecq,lee2022flexround}. Whereas, QAT learns quantized weights in the training phase or during retraining \cite{Chen_2023_ICCV,huang2023efficient,bhalgat2020lsq,esser2020learned,nagel2022overcoming}. \cite{frantar2022optimal} attempt to compress a network using both pruning and quantization. These methods, however, only focus on reducing the bit precision while maintaining the accuracy of the original model with little to no fairness control. 

\subsection{Mixed-Precision Quantization} 
Mixed-precision quantization is widely used because low-precision PTQ tends to have pitfalls in accuracy. \cite{wu2018mixed} uses neural architecture search to find suitable precisions for different layers. A mixed-precision integer-only inference for faster computation was explored by \cite{yao2021hawq}. \cite{dong2019hawq} uses Hessian spectra of the network layers to determine the precision of the layers. 

\subsection{Algorithmic Bias and Fairness} 
There are pressing concerns considering the surge of neural network-based models for everyday use. In the context of neural networks, adversarial learning methods \cite{wadsworth2018achieving,xu2021robust} are often used to achieve fairness. \cite{du2021fairness} de-bias the classification head to improve the fairness of networks. However, most fair learning algorithms suffer from tradeoffs, for e.g., \cite{zhao2022inherent} prove that one such tradeoff exists between statistical parity and model accuracy when learning fair representations. Several studies measure algorithmic bias using standard datasets \cite{amini2019uncovering} or synthetic datasets \cite{liang2023benchmarking}. These studies consider a fully trained neural network that is not compressed. However, in this work, we dive into the fairness of compressed neural networks via quantization through the lens of the changes and impacts occurring in the model due to quantization. In \cite{tran2022pruning}, the authors recognized that pruning may have a disparate impact on model accuracy and attribute it to changes in the gradient norms and eigenvalues of the Hessian matrix. In the context of quantization with an imbalanced class distribution, \cite{chen2022climbq} proposed \textit{HomoVar} loss to balance classes during quantization. Using skip-connections and Dirichlet distribution, \cite{zhou2023novel} create a framework for mixed-precision quantization to dampen disparity. 

However, these prior works do not explain the causes of the disparate impact of quantization, especially from the perspective of the impact factors inside the network. 




\section{Problem Formulation}

Consider a classification task involving a dataset $D$ with $M$ input samples $X= \{x_1, x_2, \cdots , x_i, \cdots, x_M\}$ and corresponding classes $Y=\{y_1, y_2, \cdots , y_i, \cdots , y_M\}$ where $y_i \in G$ groups (classes). The objective is to learn a classifier $f_\theta$ with parameters $\theta \in \mathbb{R}^K$, where $K$ is the number of parameters in the network. The risk function obtained by using cross-entropy as the loss function to measure the discrepancy between the predicted and actual labels under empirical risk minimization (ERM) \cite{ERM1991} is: 
\begin{equation}\label{eq:loss_definition}
    L(\theta;D) = -\frac{1}{M} \sum_{i=1}^{M} \sum_{g=1}^{G} y_{ig} \cdot \log(p_{\theta}(x_i))_g
\end{equation}
where $p_\theta(x_i) = \sigma(f_\theta(x_i))$ and $\sigma(z_{g}) = \frac{e^{z_g}}{\sum_j e^{z_j}}$.
The best solution to this optimization problem is given by, $\theta_o = \underset{\theta}{\mathrm{argmin}} \  L(\theta;D) $. Note that this definition pertains to an uncompressed model. Subsequently, let $\theta_q$ be the weights of a quantized network such that $\theta_q = T(\theta_o)$, where $T$ is a quantization function and $q$ is the number of bits used to represent the weights of the network. For example, if the network was quantized to use 8-bit representations, the network parameters are denoted by $\theta_8$. Let $\widetilde{\theta_q}$ denote the dequantized weights obtained by scaling $\theta_q$ to floating point numbers, $\widetilde{\theta_q}=S.\theta_q$, where $S$ is the set of scaling factors. As a result, the risk functions for the original and compressed models are given by $L(\theta_o;D_g)$ and $L(\widetilde{\theta_q};D_g)$, respectively.

\subsection{Fairness Analysis} 
Visualizing fairness via changes to a loss function is challenging due to its multidimensional nature. Consequently, we rely on its correlation with model accuracy and observe its changes across groups. Among the differences, the largest discrepancy can represent how unfair the model actually is. In light of this, we propose Fairness Violation Observed (\texttt{FVO}): 
\begin{equation} 
     \texttt{FVO}(\theta;D) = \underset{g,g'\in G}{\mathrm{max}} |Acc(\theta,D_g)-Acc(\theta,D_{g'})|
     \label{eq:FVO}
\end{equation}
where $Acc(\theta, D_g)$ and $Acc(\theta,D_{g'})$ represent the accuracy of groups $g$ and $g'$ for parameters $\theta \in \{\theta_o,\widetilde{\theta_q}\}$.




\paragraph{Why \texttt{FVO}?}
\texttt{FVO} is interpretable and generalizes well to multi-class tasks. Further, minimizing \texttt{FVO} inherently captures equalized odds \cite{equalized_odds} when accuracy differences arise from varying true positive and false positive rates across groups. Other advantages include:
\begin{enumerate}
\item FVO is not limited to a specific pair of groups and applies to the overall set of groups involved. 
\item It directly relates to accuracy, which is an interpretable metric for computer vision tasks.
\item It is practical for situations seeking to balance model performance with fairness 
\end{enumerate}

\subsection{What is the Best Model?} 
A model with low \texttt{FVO} indicates that the performances across all groups are similar. However, it does not guarantee that the model performs well overall. That is because, in particular, when all the groups have low accuracies, the model may end up with worse overall performance. In order to choose a model that is both fair and accurate, we consider both \texttt{FVO} and the overall accuracy, \texttt{OA}, together. Thus, when comparing different approaches for fairness, our preference aligns with the model that maximizes \texttt{OA} and minimizes \texttt{FVO}:  
\begin{equation} \label{eq:best_model_criteria}
     \underset{\texttt{OA}} {\mathrm{\max}} \ \underset{\texttt{FVO}}{\mathrm{\min}} \ f_{\theta,D}
\end{equation}

\paragraph{Setup} For the investigations presented in this paper, we use per-tensor uniform post-training quantization (PTQ) \cite{nagel2021white} for weights, based on the implementation in \cite{banner2019posttraining} for integer quantization. In particular, for $\mathtt{fp16}$ experiments, we used half-precision computation from the PyTorch library. Note that the integer weights are scaled to floating points during inference. The following experiments are on UTKFace dataset \cite{zhifei2017cvpr} with the task of classifying the ethnicity using a ResNet18 architecture, where the weights are quantized to 16, 8, 4, and 2 bits. The original network's precision is 32 bits. 

\begin{figure*}[th!]
    \centering
    \captionsetup{justification=centering}
    \includegraphics[width=0.95\linewidth]{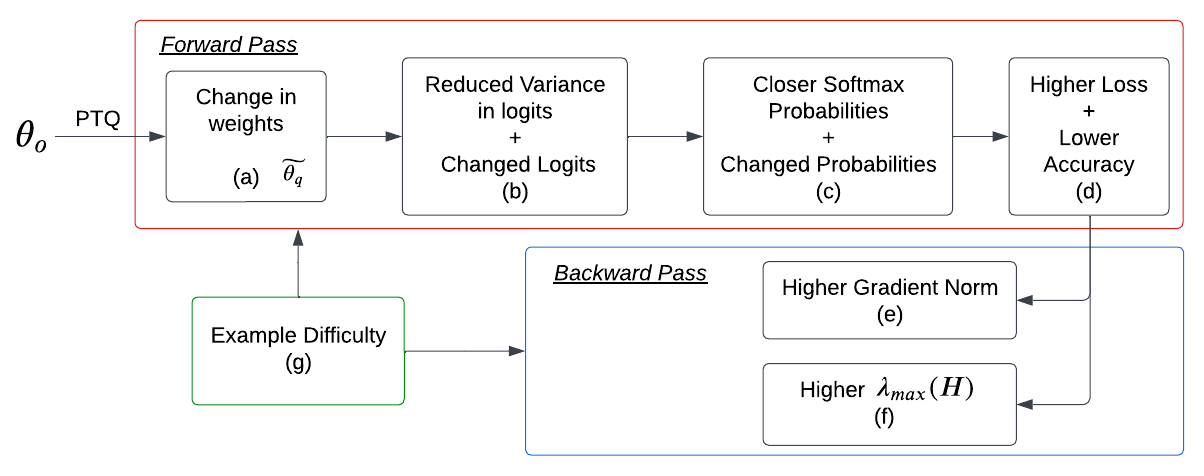}
    \caption{The impact flow of quantization.}
    \label{fig:quant-flow}
\end{figure*}

\section{Factors Impacting Fairness}


The impact of quantization occurs through multiple stages, as shown in Fig.~\ref{fig:quant-flow}. During the forward pass, the effect of the changes in weights propagates throughout the network and leads to changes in logits, whose behavior is reflected in the softmax probabilities and, therefore, the loss. To better understand and visualize the effects of higher loss on the network weights, we use backpropagation without actually updating the weights, motivated by the second order Taylor Series expansion of the loss function at point $x_c$,
\begin{equation}
    L(x) = L(x_c) + \nabla L(x_c) \cdot (x - x_c) + \frac{1}{2} (x - x_c)^T H (x - x_c)
\end{equation}
Here, $\nabla L$ represents the gradient $G$. Now, for every group and precision, we study the gradient norm $||G_g^L||$ and the largest eigenvalue of the Hessian matrix $\lambda_{max}(H_g^L)$ for the loss function $L$. The gradient norm helps us understand how far away the solution is from a better state in the solution space. Whereas, the eigenvalues of the Hessian matrix provide crucial information about the steepness in the loss surface. Quoting from \cite{tran2022pruning}, the maximum of the eigenvalues indicates how well the solution can separate the groups. \cite{keskar2017largebatch,li2018visualizing} support that the top eigenvalues of the Hessian matrix aid in understanding the loss landscape. We next look at each stage of the impact flow in detail.

\begin{figure}
    \centering
    \captionsetup{justification=centering}
    \includegraphics[width=0.45\linewidth]{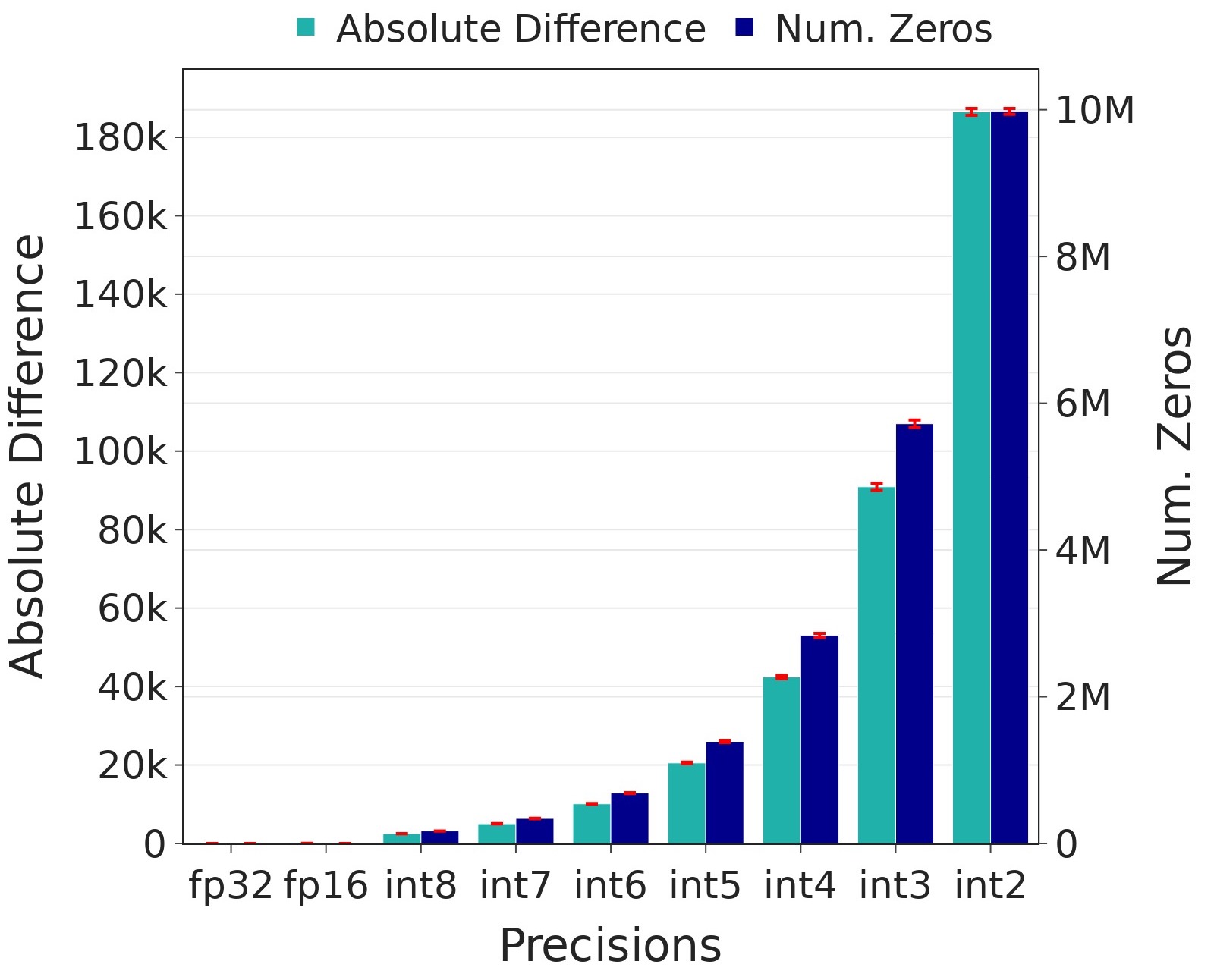}
    \caption{Changes in weights due to quantization}
    \label{fig:changes_in_weights}
\end{figure}

\subsection{Changes in the Weights}
The root cause of the impact flow of quantization is the change in weights of the network. The absolute difference in the weights is defined as 
\begin{equation}
   \mbox{Absolute difference in the weights} = \sum_{k=1}^K |\widetilde{\theta}_{q,k}-\theta_{o,k}|
\end{equation}
However, the impact does not only include the absolute difference, but also involves the fraction of ``zero'' weights induced by quantization. While the former quantifies how much the weights have deviated from the original values, the latter is indicative of the loss of information due to sparsity defined by 
\begin{equation}
    \mbox{Sparsity} = \frac{1}{K}{\sum_{k=1}^{K} I(\theta_{.,k} = 0)}
\end{equation}
where $I$ denotes the indicator function, $\theta_. \in \{ \theta_o,\theta_q\}$. The absolute difference is controlled by the reduction in the precision of the weights. For example, $\theta_4$ has 28 lesser bits to represent the weights in comparison to $\theta_{32}$, which persists even after scaling by $S$. Whereas, sparsity increases when the weights are pushed to the '$0$ bin' during quantization which continues to remain as $0$s even after scaling. While achieving higher compression, this effect is similar to (unstructured weight) magnitude pruning \cite{han2015learning,zhu2017prune,Frankle2019ICLR}, where some of the weights of the network are changed to $0$.

\begin{figure} 
\centering
    \begin{subfigure}{0.37\columnwidth}
        \captionsetup{justification=centering}
        \includegraphics[width=0.85\linewidth]{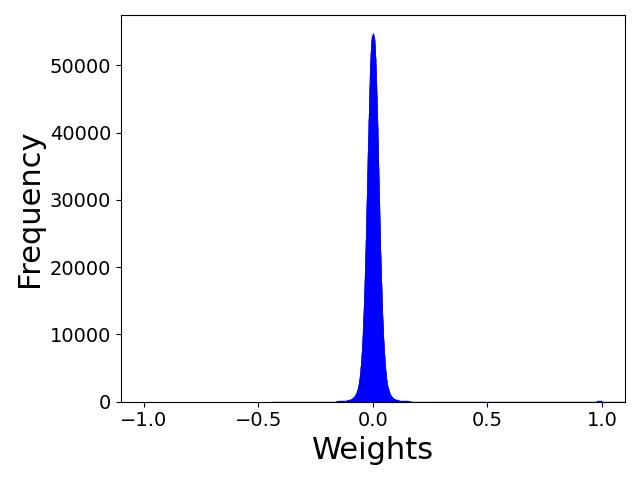}
        \caption{\texttt{fp32}}
        \label{fig:fp32_dist}
    \end{subfigure}
    ~
    \begin{subfigure}{0.37\columnwidth}
        \captionsetup{justification=centering}
        \includegraphics[width=0.85\linewidth]{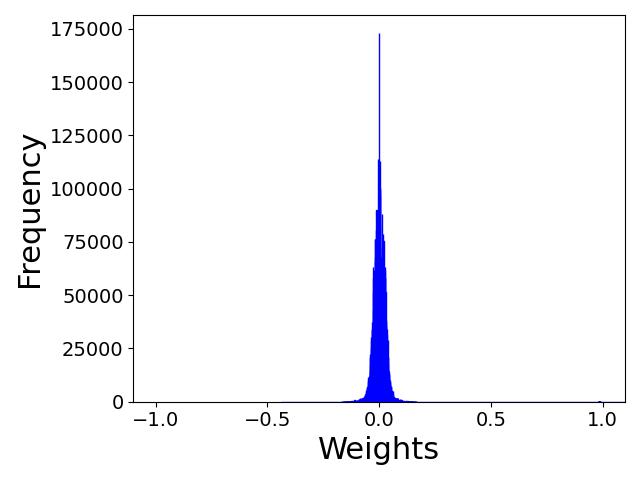}
        \caption{\texttt{int8}}
        \label{fig:int8_dist}
    \end{subfigure}
    ~
    \begin{subfigure}{0.37\columnwidth}
        \captionsetup{justification=centering}
        \includegraphics[width=0.85\linewidth]{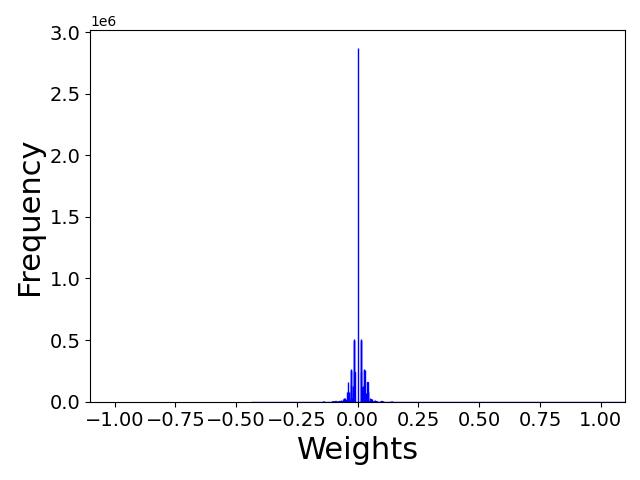}
        \caption{\texttt{int4}}
        \label{fig:int4_dist}
    \end{subfigure}
    ~
    \begin{subfigure}{0.37\columnwidth}
        \captionsetup{justification=centering}
        \includegraphics[width=0.85\linewidth]{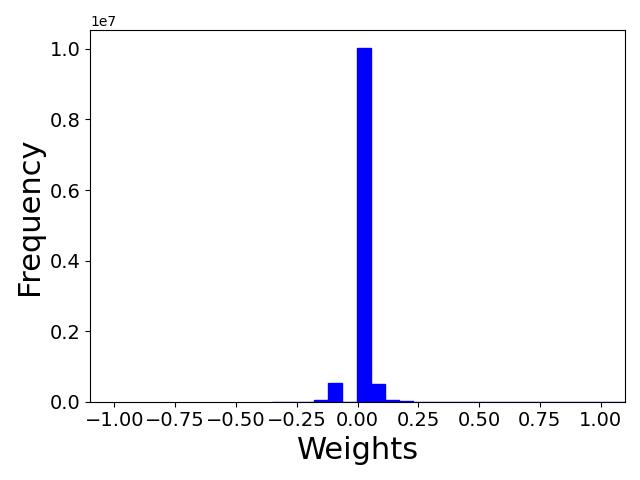}
        \caption{\texttt{int2}}
        \label{fig:int2_dist}
    \end{subfigure}
    \caption{Weight Distributions}
    \label{fig:weight-distributions}
\end{figure}

Fig.~\ref{fig:changes_in_weights} illustrates an increase in both absolute difference in weights and sparsity as precision reduces. On the other hand, Fig.~\ref{fig:weight-distributions} shows the weight distribution of $\widetilde{\theta_q}$ for different precisions, indicating a distribution shift towards the center with reducing precision. Clearly, reducing the precision increases the sparsity of the network, therefore, making it more like a pruned network (by weight magnitude). As shown in \cite{tran2022pruning}, increasing the pruning ratio, i.e., increasing sparsity, has a disparate impact on the accuracy of the model. With higher sparsity in a quantized model, disparity worsens, analogous to the impact of sparsity in a pruned model. These combined changes affect the logits of the network, which we analyze next.


\begin{figure*} [htb!]
\centering
    \begin{subfigure}[t]{0.99\linewidth}
    \centering
        \includegraphics[width=\linewidth]{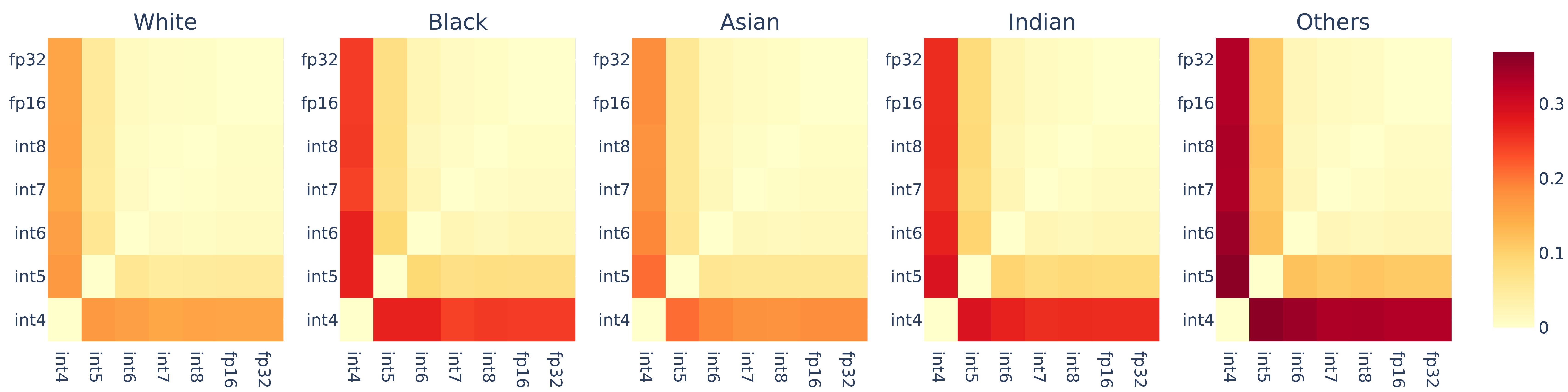}
        \caption{Cosine Distance between logits. Darker shade implies higher distance.}
        \label{fig:logits_distance}
    \end{subfigure}

    \begin{subfigure}[t]{0.99\linewidth}
        \centering
        \includegraphics[width=1\textwidth]{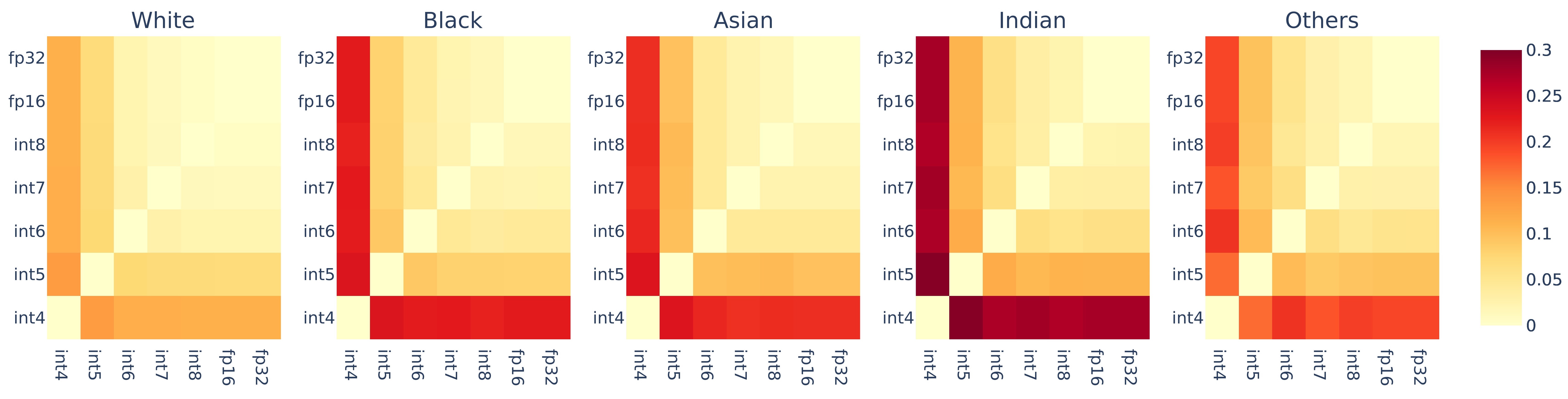}
        \caption{L1 Distance between logits for different quantization levels across classes - UTKFace. Note that L1 does not capture the dissimilarity as finely as cosine distance.} 
        \label{fig:utkface_logit_cl1}
    \end{subfigure}
    
    \begin{subfigure}[t]{0.99\linewidth}
        \centering
        \includegraphics[width=1\textwidth]{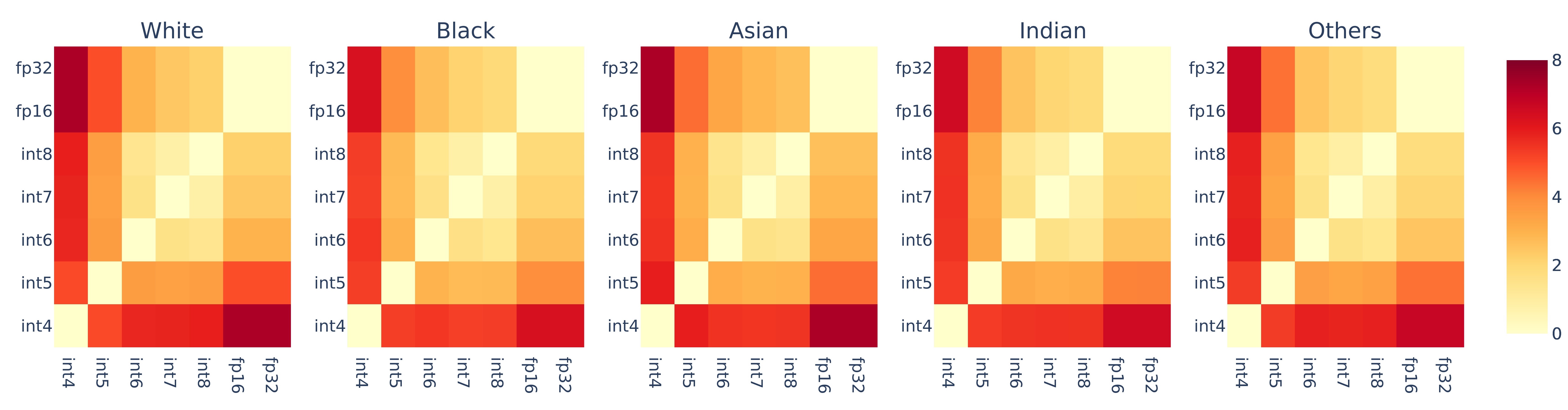}
        \caption{L2 Distance between logits for different quantization levels across classes - UTKFace. Note that L2 is worse in comparison to both L1 and CD, therefore, indicating that the shift due to quantization is better observed through an angle-based measure than a norm-based measure.} 
        \label{fig:utkface_logit_l2}
    \end{subfigure}

    \begin{subfigure}[t]{0.75\linewidth}
    \centering
        \includegraphics[width=\linewidth]{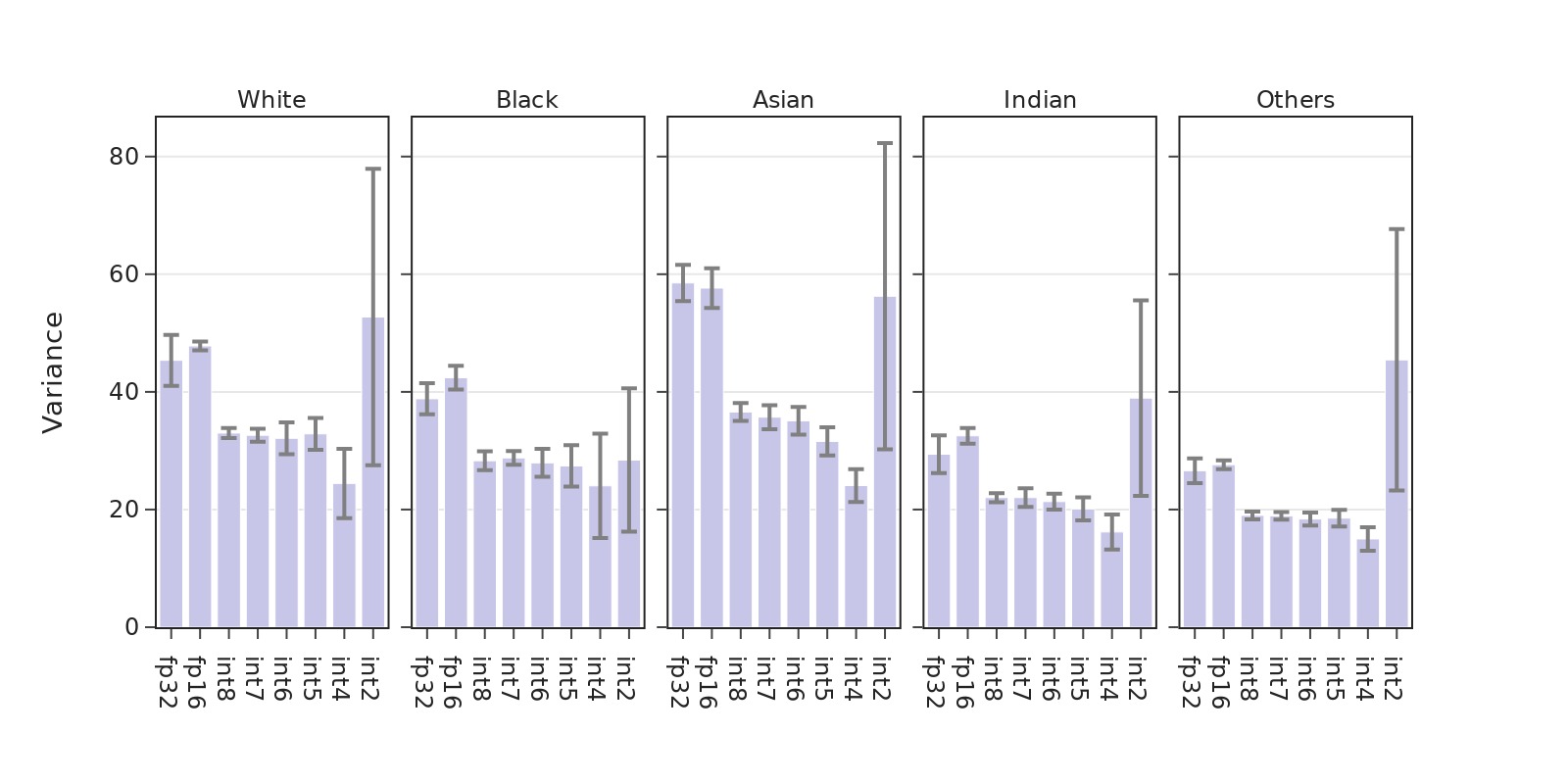}
        \caption{Decrease in precision leads to reduction in variance in logits}
        \label{fig:logit-variances}
    \end{subfigure}
    \caption{Logits analysis}
\end{figure*}





\begin{figure*} [t]
\centering
    \begin{subfigure}[t]{0.85\textwidth}
    \centering
        \includegraphics[width=\linewidth]{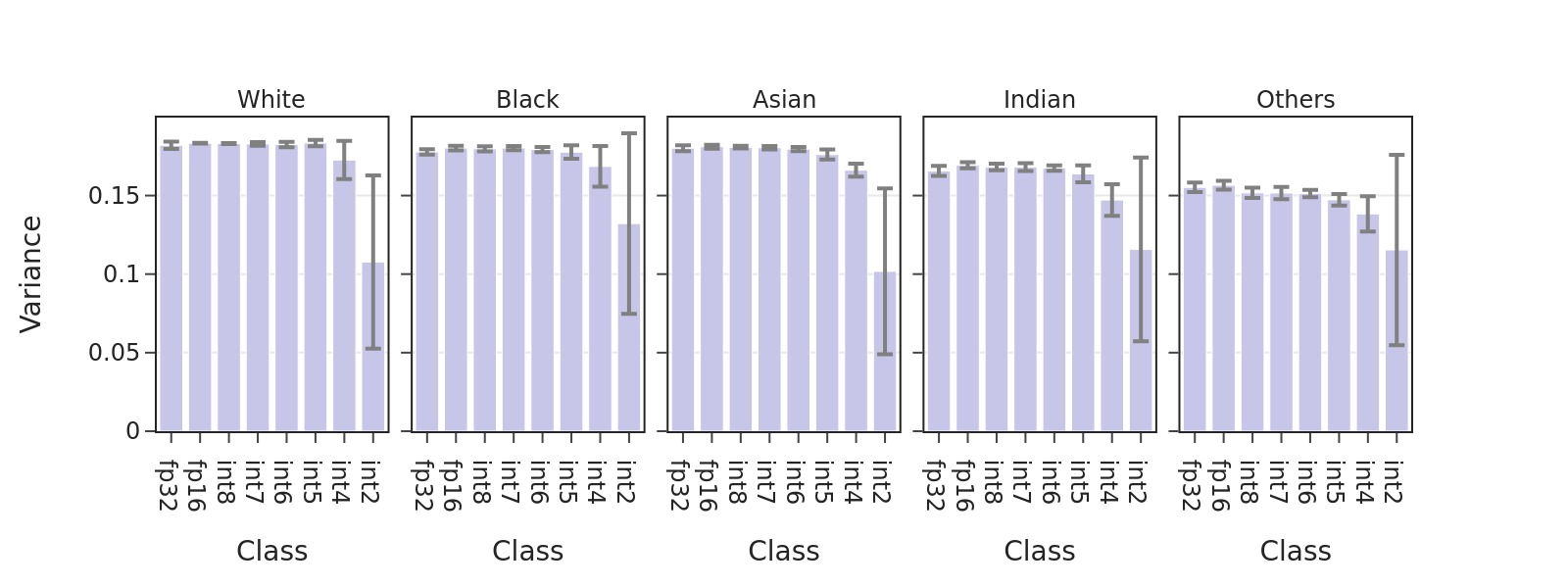}
        \caption{Reduced variance in logits has a persistent presence in softmax probabilities}
        \label{fig:softmax-variances}
    \end{subfigure}%
    
    \begin{subfigure}[t]{0.8\linewidth}
    \centering
        \includegraphics[width=\linewidth]{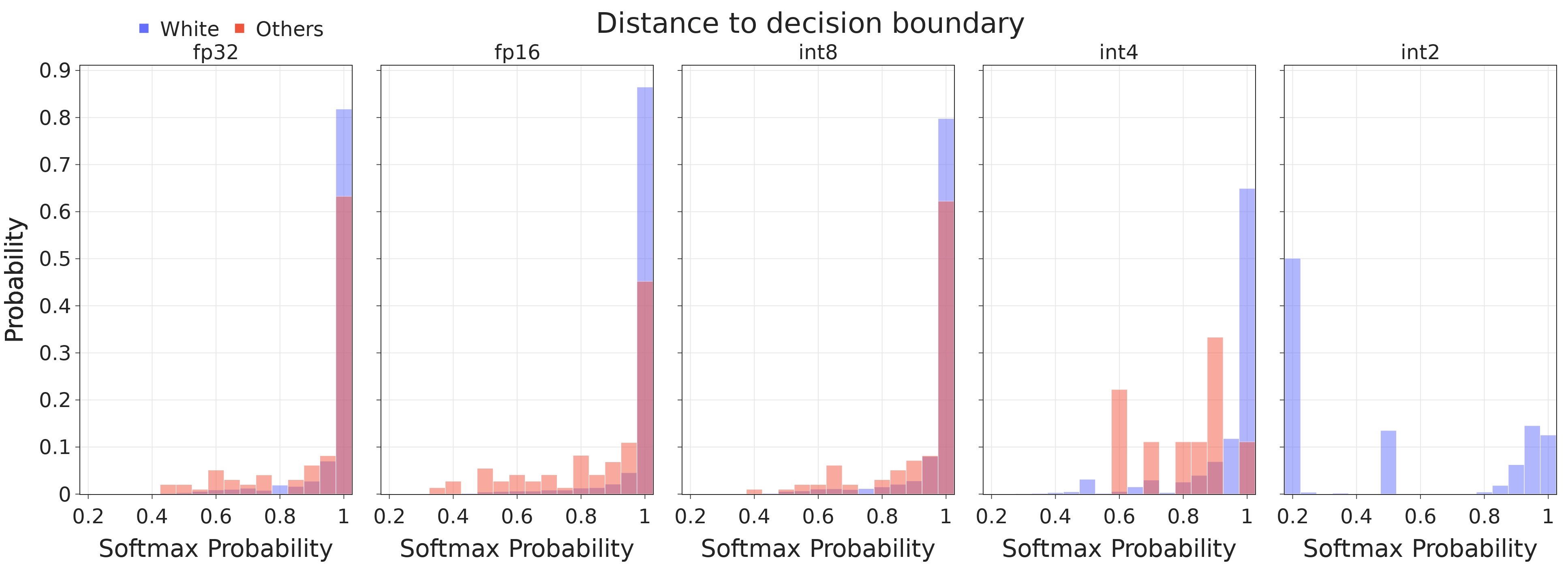}
        \caption{The probability distribution of the distance to decision boundary (softmax probability). Notice the distribution shift of \texttt{Others} to the left and eventually disappears as precision reduces.}
        \label{fig:dodb_distribution}
    \end{subfigure}
    \caption{The effect on Softmax probabilities}
\end{figure*}


\subsection{The Effect on Logits and Probabilities}


The logits undergo two transformations as a result of quantization. First, the numerical values undergo a significant change, causing distinct differences in the magnitudes and resulting in different highest values among the logits. At lower precisions, this shift can cause the highest values to occupy incorrect positions in the logit vector, consequently leading to inaccurate predictions. Second, the variance between the logits is affected, resulting in different loss values.

We study the change in numerical values using cosine distance, defined as,
\begin{equation}
    CD(A,B) = 1 - \frac{{A \cdot B}}{{\|A\| \cdot \|B\|}}
\end{equation}
where $A$ and $B$ are two vectors of equal length. Let the average cosine distance between $f_{\widetilde{\theta_q}}$ and $f_{\theta_o}$ across the samples of a group be, 
\begin{equation}
    \mbox{Average cosine distance} = \frac{1}{|G|}\sum_i^{|G|} CD (f_{\widetilde{\theta_q}}(x_i),f_{\theta_o}(x_i))
\end{equation}
Fig.~\ref{fig:logits_distance} shows that the angle between different quantization levels is largest for the minority class \texttt{Others} and the least for the majority class \texttt{White}. CD captures the changes that occur in the logits due to quantization, however, norm based metrics fail to capture it, as observed in Fig.~\ref{fig:utkface_logit_cl1} and Fig.~\ref{fig:utkface_logit_l2}. Note that we are not able to show $\theta_2$ and $\theta_3$ as they produce null vector logits for some images which makes cosine distance inapplicable. 

The mean variance among logits within each group, represented as, 
\begin{equation}
    \mbox{Mean variance of logits} = \frac{1}{|G|}\sum_i^{|G|} \text{Var}(f_\theta(x_i))
\end{equation}
decreases with decreasing precision, as observed in Figure \ref{fig:logit-variances}. Notably, the group \texttt{White} exhibits the highest variance, while the \texttt{Others} group demonstrates the least variance. This reduction indicates that the separability of groups worsened due to quantization.

These changes in logits subsequently induce both a reduction in variance and a distribution difference in the softmax probabilities. At lower precisions, there is a substantial decrease in variance across all groups, with the \texttt{Others} group being affected the most, as illustrated in Fig.~\ref{fig:softmax-variances}. This reduced variance is analogous to the output-softening nature of the high-temperature scaling, which softens the logits of the network. Further, the disruption in the softmax probability distribution links to the inability of the precision to capture the original model's behavior. The softmax probability can be viewed as a Distance To the Decision Boundary ($DTDB$). We define $DTDB_{i,g}$ as the softmax probability obtained for each sample $i$ belonging to group $g$, and that is plotted in Fig.~\ref{fig:dodb_distribution}. If $DTDB_{i,g} > DTDB_{i,g'}$, then group $g$ is farther away from the decision boundary than $g'$, which implies an easier classification. Fig.~\ref{fig:dodb_distribution} shows a strong leftward shift of distribution for \texttt{Others} unlike \texttt{White}, indicating that reduced precision induces uncertainty in the model for minority classes.

 
\begin{figure}
    \centering
    \captionsetup{justification=centering}
    \includegraphics[width=0.8\columnwidth]{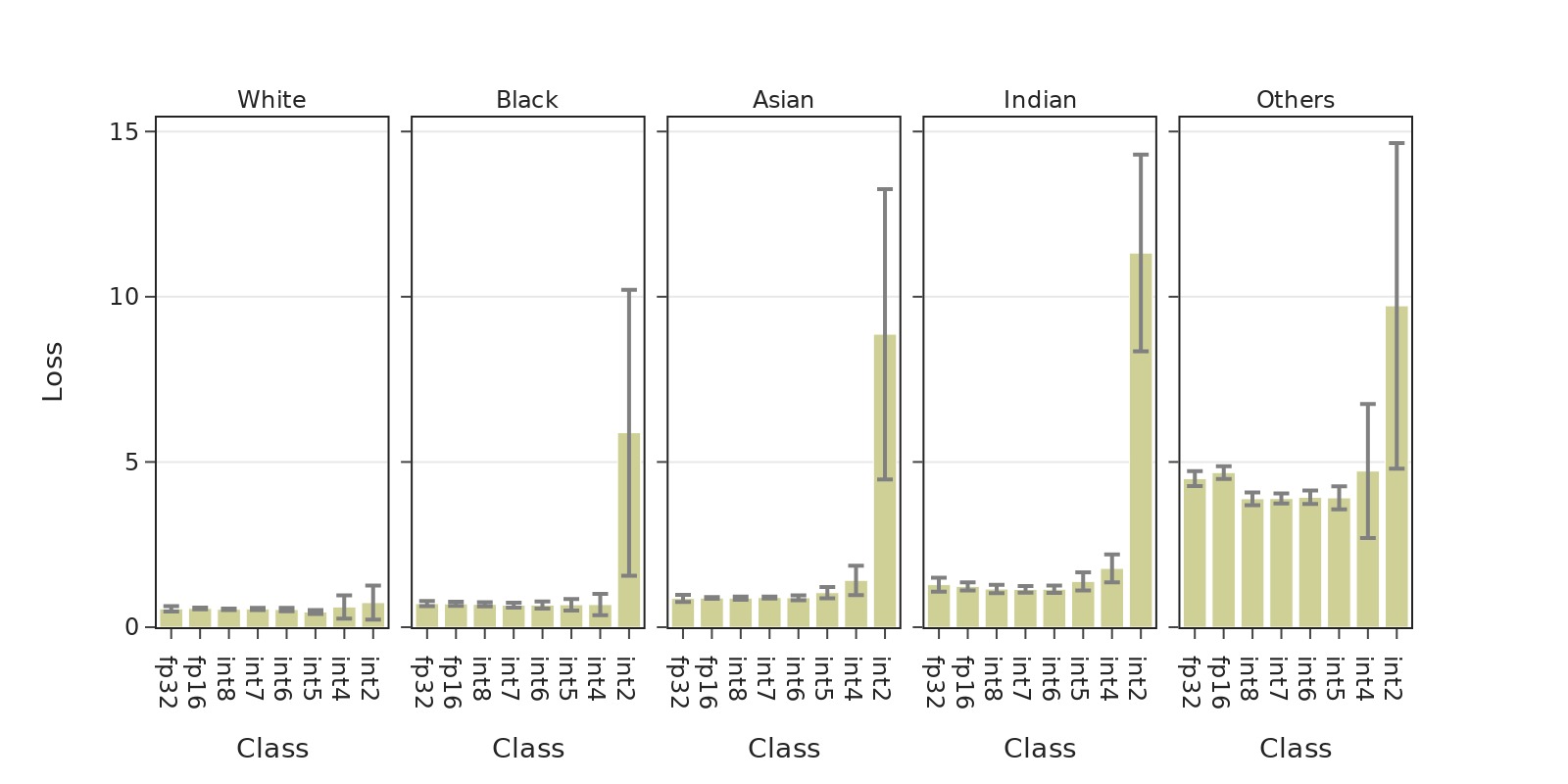}
    \caption{Higher group loss for \texttt{Others} due to PTQ}
    \label{fig:group_losses}
\end{figure}

\begin{figure}[b]
\centering
    \begin{subfigure}[t]{0.4\textwidth}
    \centering
        \captionsetup{justification=centering}
        \includegraphics[width=0.7\linewidth]{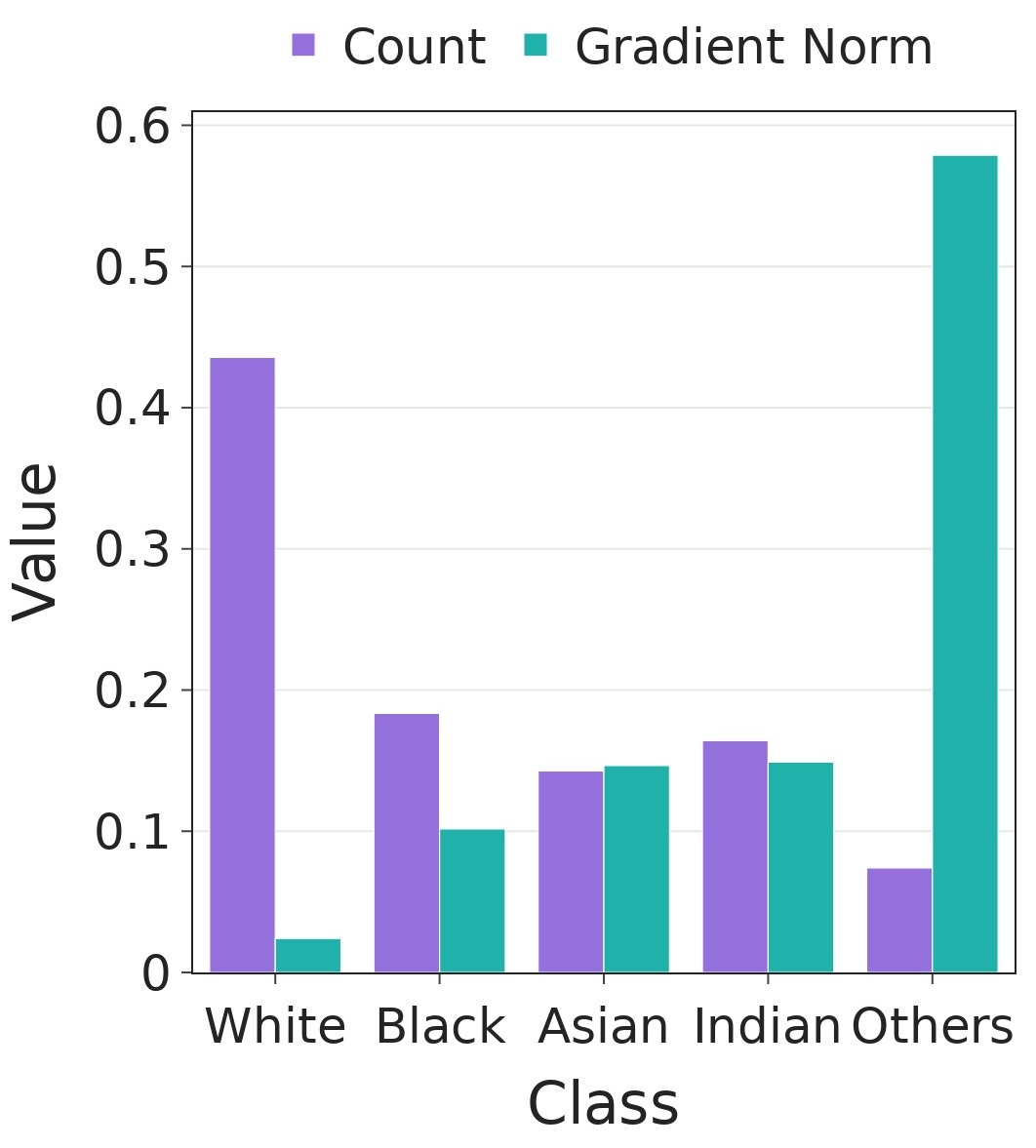}
        \caption{Gradient norm vs. group size}
        \label{fig:grad_norm_group_size}
    \end{subfigure}
    \quad \quad
    \begin{subfigure}[t]{0.4\textwidth}
    \centering
        \captionsetup{justification=centering}
        \includegraphics[width=0.7\linewidth]{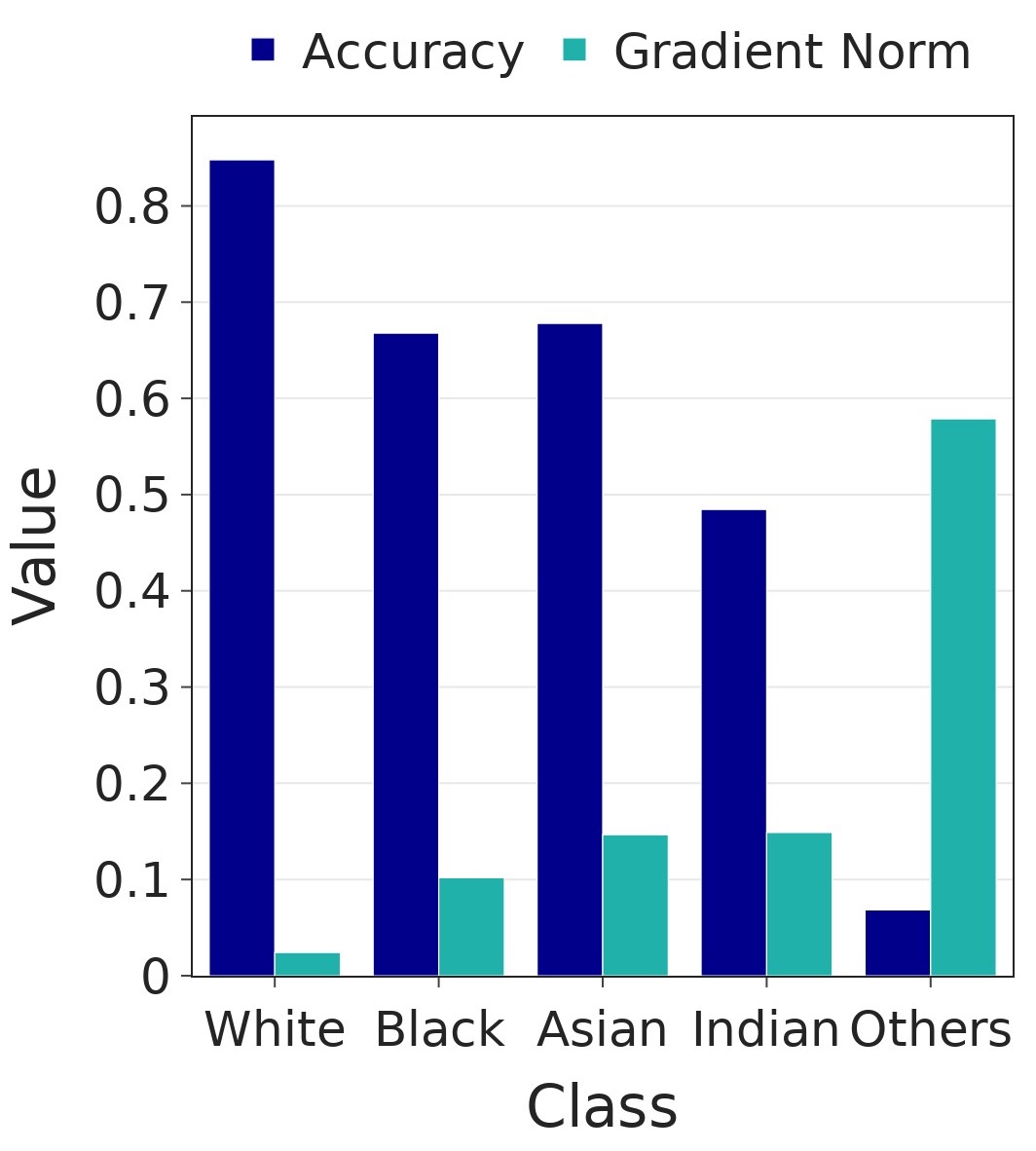}
        \caption{Gradient norm vs. accuracy}
        \label{fig:grad_norm_accuracy}
    \end{subfigure}
    \caption{Trends of gradients against group size (normalized) and accuracy on an Int4-quantized model for ResNet18}
\end{figure}

\begin{figure} [t]
 \centering
    \begin{subfigure}[t]{0.48\textwidth}
    \centering
    \captionsetup{justification=centering}
        \includegraphics[width=0.7\linewidth]{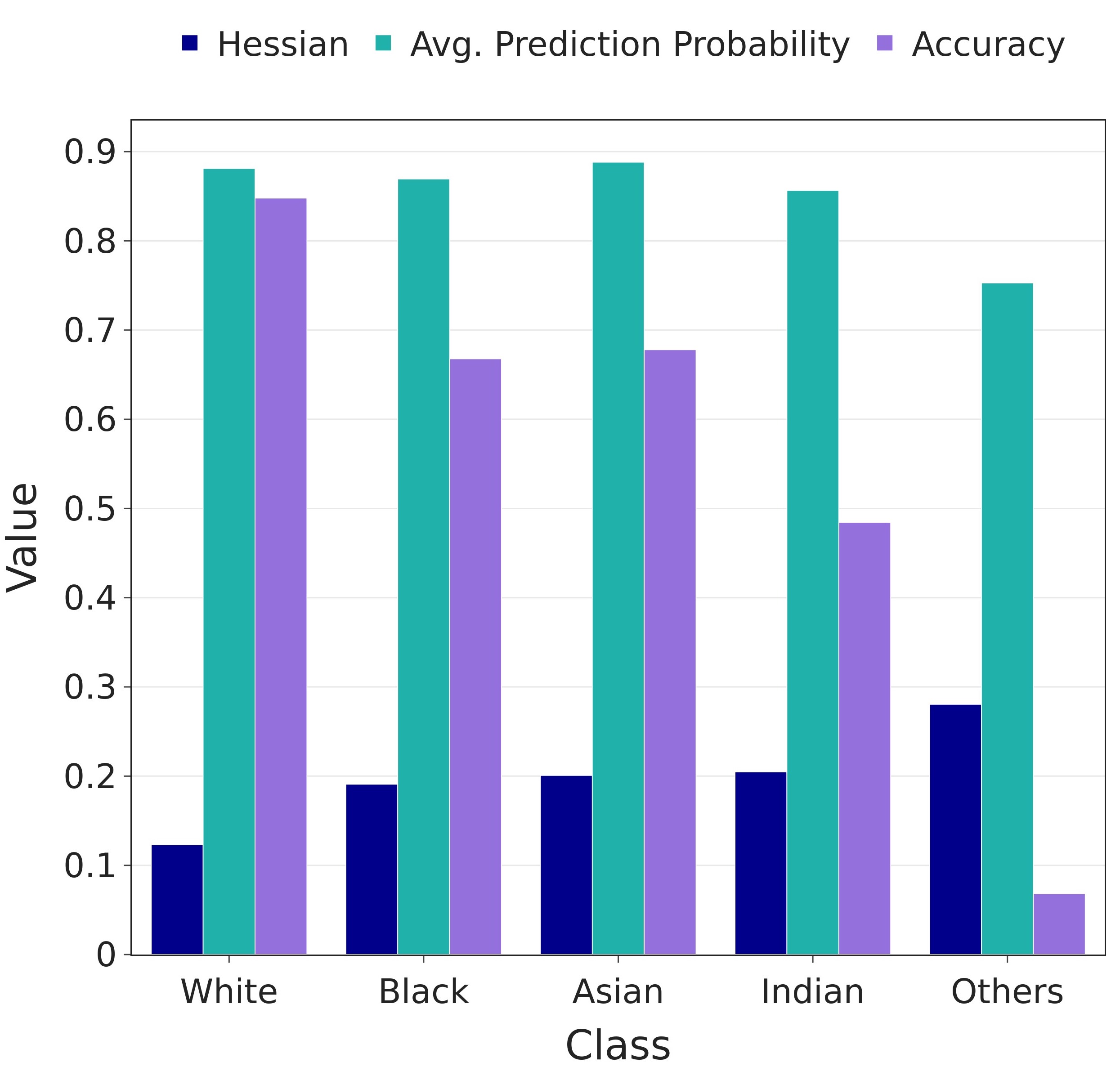}
        \caption{The largest Hessian eigenvalue ($\lambda_{max}$) is inverse to accuracy and average prediction probability}
        \label{fig:hessian_acc_dtdb}
    \end{subfigure}
    \quad 
    \begin{subfigure}[t]{0.48\textwidth}
    \centering
    \captionsetup{justification=centering}
        \includegraphics[width=0.7\linewidth]{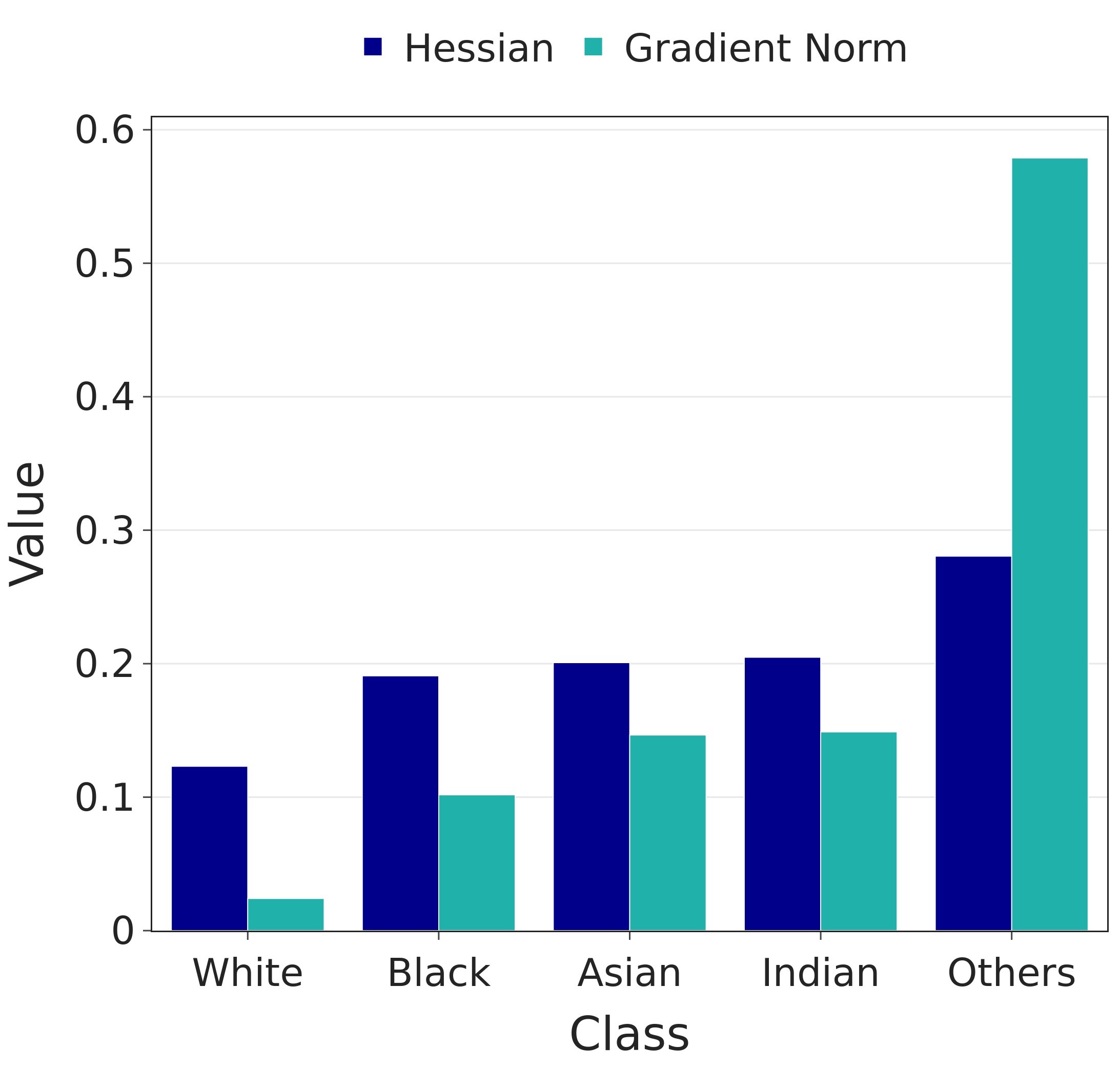}
        \caption{The largest Hessian eigenvalue ($\lambda_{max}$) aligns with the gradient norm}
        \label{fig:hessian_grad_norm}
    \end{subfigure}
    \caption{The largest Hessian eigenvalue for different groups}
\end{figure}

\subsection{Contribution to Loss and Accuracy}
The reduced variance in softmax probabilities, together with the changed values, adversely affect the loss and accuracy of the model as depicted in Fig.~\ref{fig:group_losses}. The per-group loss is highest for \texttt{Others} and least for \texttt{White}. In addition, it is reflected as a direct impact on the accuracy of the model, as observed in Fig.~\ref{fig:base_accuracies}. These circumstances indicate a clear, unfavorable movement of the model in the optimization space for all the classes, due to quantization, with the most affected being \texttt{Others}. In order to better understand this degraded position, we backpropagate the loss and observe how the gradient norm and Hessian are affected.

\subsection{Observing Unfairness through Gradient norms}
The gradient norm provides insight into the convergence of the optimization problem, indicating the proximity of the solution in the optimization space to a local minima \cite{zhao2022penalizing}. We find the  group gradient norm for a quantized network using the gradients obtained by passing the test set (without weight updates) and evaluating the $\ell_2$ norm, given by, 
\begin{equation}
    G(\widetilde{\theta_q}; D_g) = \sqrt{\sum_{k=1}^{K} \left(\frac{\partial L(\widetilde{\theta_q}; D_g)}{\partial \widetilde{\theta_{q,k}}}\right)^2}
\end{equation}

This measure also signifies the extent of gradient updates necessary for the model to improve its prediction. Next, we examine factors that exhibit a correlation with the gradient norm.


Consider the situation when $D$ is passed as a single batch for gradient updates. Initially, the averaged gradients are dominated by classes with a higher number of samples. This effect persists even when there are mini batches, although lower in impact. Therefore, the gradients are also controlled by the class distributions and batch size. In addition, the initial gradients are heavily dependent on the initialization of $\theta$. However, the effects of batch size (if moderate) and initialization dampen as the network trains further. We therefore look at the effects of per-group sample counts of the test set on the gradient norm. Fig. \ref{fig:grad_norm_group_size} shows an inverse trend between the gradient norm and group sizes for $\theta_4$. Notice the huge disparity between the Gradient norm of \texttt{White} and \texttt{Others}. We argue that this occurs due to the initial dominance of the majority classes in the gradients which are inverted at some point during training and are reflected post training and quantization. It further reflects an inverse trend with the accuracy of the model as observed in Fig.~\ref{fig:grad_norm_accuracy}.





\subsection{Reflection of unfairness on the Hessian}

$\lambda_{max}(H_g^l)$ helps explain the steepness of the loss surface at that point in the solution space for a particular group. Fig.~\ref{fig:hessian_acc_dtdb} shows that $\lambda_{max}$ and accuracy move in opposite directions, indicating a larger $\lambda_{max}$ for the minority group. This implies that the steepness is the highest for \texttt{Others}, and a corresponding update to the weight would cause a higher reduction in the loss as compared to any other group. 
To capture the average of the highest softmax prediction probabilities across the groups, we define \emph{average prediction probability}, 
\begin{equation}
    \mbox{Avg. prediction prob.} = \frac{1}{|G|}\sum_i^{|G|} max(\sigma(f_\theta(x_i))) 
\end{equation}
We also observe in Fig.~\ref{fig:hessian_acc_dtdb} that the average prediction probability is lowest for the group with the highest $\lambda_{max}$ and vice versa. Fig~\ref{fig:hessian_grad_norm} shows gradient norm and $\lambda_{max}$ moving toward the same direction, indicating that quantization induces a combined effect on them. 






\subsubsection{Comparisons for Different Quantization Precisions}
\begin{figure}[t]
    \centering
    \captionsetup{justification=centering}
    \includegraphics[width=0.45\linewidth]{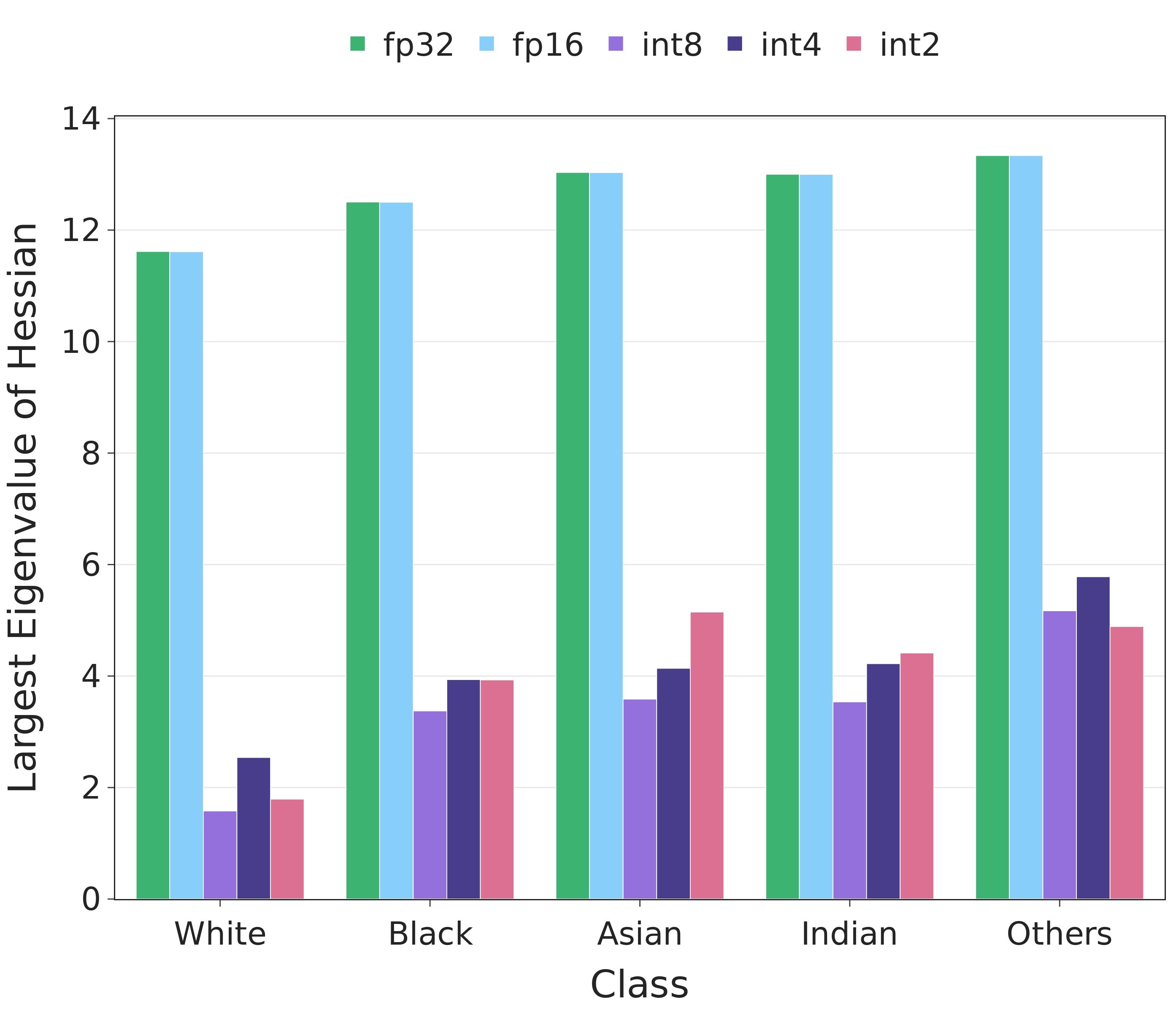}
    \caption{$\log(\lambda_{max})$ for different groups and precisions }
    \label{fig:hessian_across_qtypes}
\end{figure}

\noindent The trend across groups is similar for all quantization levels (same color across different groups) in Fig. \ref{fig:hessian_across_qtypes}. However, for different quantization levels within a group, we observe that for $\mathtt{fp32}$ and $\mathtt{fp16}$, $\log(\lambda_{max})$ is almost equal, but for the integer precisions, the overall trend is increasing. One would expect that when we quantize with lower bits, the ability to be closer to the original weight degrades and the solution ends up at an inferior point in space. However, it need not be true that the steepness of the inferior point is bad too. Fig.~\ref{fig:hessian_across_qtypes} therefore indicates that the scope for improvement is close to this order, $\mathtt{int2} > \mathtt{int4} > \mathtt{int8}$, in most cases.

\subsection{Example Difficulty} \label{example-difficulty}

When sub-sampling for majority classes to create a balanced dataset, we observed that during training, the per-group accuracy was initially lower for the group \texttt{Others}, compared to the other classes. The model then reaches equal training accuracy for every class towards the end, yet performs poorly on the test set for \texttt{Others}. Given the scenario that the number of images used to train is nearly the same, the above two observations explain that the data in the test set for \texttt{Others} is rather difficult for the network to classify correctly. We attribute this lack of generalizability to the relatively complex facial features in the \texttt{Others} class, such as color, age, hair, and structure, which contribute to example difficulty. Example difficulty, however, has an impact during both forward and backward passes. In the forward pass, the network has already been trained better for the lower difficulty samples, which in addition, during the back pass, has a continuing effect.





    


%

\section{Mitigation Solutions: Fair Quantization}

In this section, we look at what improves the fairness of a quantized model. We first begin with methods that make an $\mathtt{fp32}$ model fairer, followed by using \texttt{QAT} as a possible solution to quantization's disparity. Finally we combine the former and latter to achieve the same with lesser data. The results performed against the UTKFace dataset are presented in this section. Please refer to the Appendix for the results of other datasets.

\begin{figure}[t]
    \centering
    \includegraphics[width=0.35\linewidth]{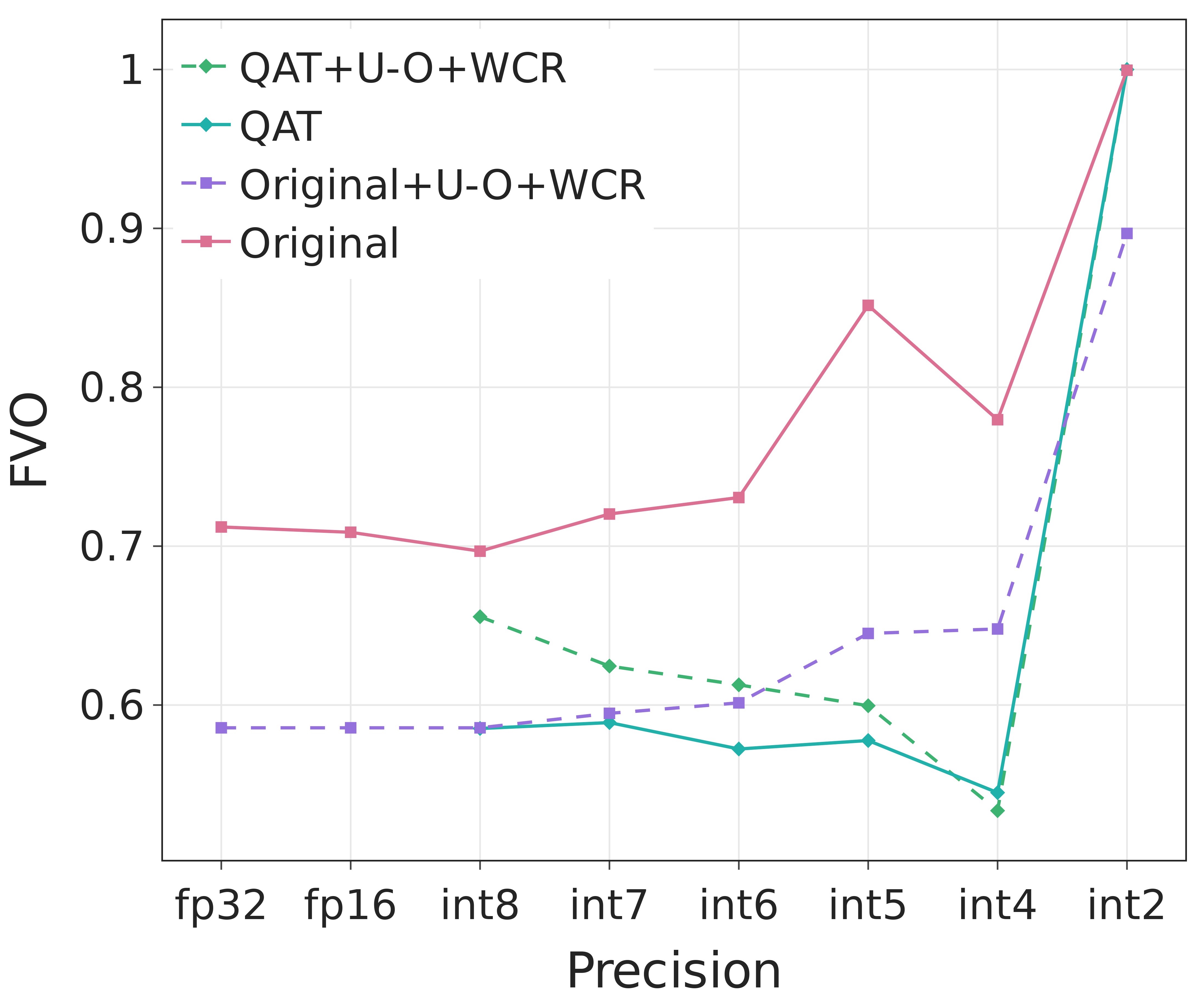}
    \caption{Comparison between mitigation solutions for Resnet18.}
    \label{fig:mitig_comparison_FVO}
\end{figure}

\subsection{A Fairer Base Model}
 Under and over-sampling (\texttt{U-O}) the majority and minority classes, respectively, reduces the per-group differences in the gradient norm during training. However, capturing the example difficulty from Sec.~\ref{example-difficulty} involves weighting the harder classes higher than the rest, so that the solution is not biased only towards getting the easier samples right, during the beginning of the optimization. Let \texttt{WCR} denote the weighted cross entropy loss function given by, 
\begin{equation}\label{eq:weighted_ce}
    L(\theta;D) = -\frac{1}{M} \sum_{i=1}^{M} \sum_{g=1}^{G} a_g \cdot y_{ig} \cdot \log(p_{\theta}(x_i))_g
\end{equation}
where $a_g$ is the weight for group g. We choose \texttt{WCR} with weights $([0.1,0.1,0.1,0.1,0.6])$ such that \texttt{Others} has the highest weight and the rest of the classes are given equal weight. Next, we train a fairer model $\theta_f$ using both \texttt{U-O} and \texttt{WCR}.

\subsection{Mixed-precision QAT}
In \texttt{PTQ}, there is no retraining to allow for the weights to be re-adjusted to the change in environment. However, in \texttt{QAT}, the losses are calculated and the weights are updated in accordance with the changing weights and activations. This provides an opportunity to control the changes in group accuracies, unlike \texttt{PTQ}. We take into account the possible oscillations that occur in weights during \texttt{QAT} and adopt the oscillation dampening method in \cite{nagel2022overcoming} to avoid it. We also use mixed precision (MPQAT) for ResNet18, where the first and last layers use $8$ bits and the activations are $32$ bits according to \cite{bhalgat2020lsq}. $L(\widetilde{\theta_q};D)$ is minimized as a result of QAT. This dampens the adverse effects of quantizing both weights without training, together with lowering the information loss in layers that are critical for classification.

\subsection{Fair QAT}
\begin{figure}[t]
    \centering
    \includegraphics[width=0.4\linewidth]{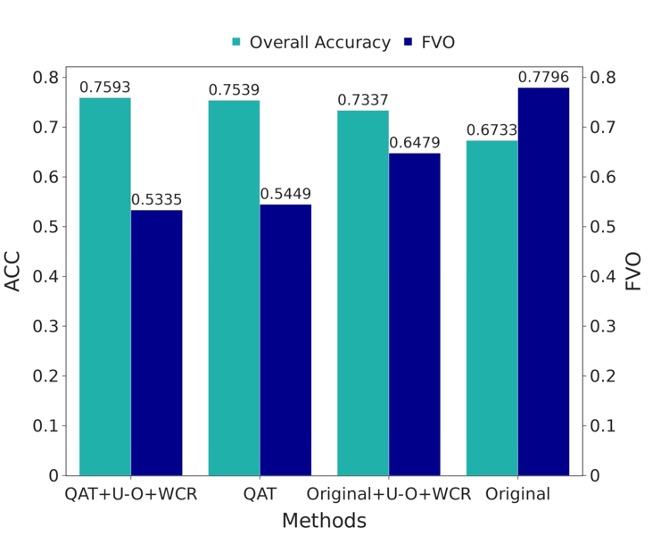}
    \caption{Tradeoffs of overall accuracy vs. \texttt{FVO} for \texttt{int4}.}
    \label{fig:mitig_comparison_FVO_acc}
\end{figure}

Combining \texttt{U-O}, \texttt{WCR}, and \texttt{MPQAT} reduces the disparate impact of Quantization for ResNet18, as observed in Fig.~\ref{fig:mitig_comparison_FVO}. In fact, \texttt{MPQAT} alone also reduces \texttt{FVO}. Here, \texttt{PTQ} has acceptable \texttt{FVO} for higher precisions, provided the original model is relatively fair. But, for lower precisions, that does not hold. We next visualize \texttt{OA} and \texttt{FVO} together in Fig.~\ref{fig:mitig_comparison_FVO_acc} showing that our method achieves both the highest \texttt{OA} and lowest \texttt{FVO}. This would provide an informed decision for the potential users to make in consideration of the tradeoffs between overall accuracy and fairness.





\section{Discussion}

\textbf{Other aspects of quantization. } In our analysis, the focus was solely on weight quantization without activation quantization. Since activation quantization further worsens the network accuracy, the results would either resemble that of weight quantization or expand the disparity that we observed. For many quantization precisions, QAT has lower FVO than the Fair QAT, which shows that QAT alone is sufficient to mitigate the disparate impact. The best hyperparameters for QAT were known for \texttt{int4}, but were unknown for the other precisions. We therefore used the same (that of \texttt{int4}) hyperparameters for all precisions. However, this should not be interpreted as refraining from the potential of other hyperparameters. The aforementioned \texttt{WCR} weights were chosen as a result of the example difficulty observed for the \texttt{Others} class. The weights are consistent for the rest of the classes, as the difficulties are nearly equal.\\







\section{Conclusion}
The disparate impact caused by PTQ is explained by an impact flow that passes across stages in the forward pass, whose effects can be visualized as a shift of the model to a sub-optimal state in the optimization landscape, using gradient norms and eigenvalues of the Hessian matrix. However, we show that utilizing simple methods, such as undersampling, oversampling, and adjusting weights in \texttt{WCR}, leads to fairer models. When QAT is used in conjunction, it gives a class of fairer quantized models with a little compromise in overall accuracy.




\bibliography{references}
\bibliographystyle{iclr2026_conference}


\end{document}